\documentclass[sigconf]{acmart}

\usepackage{multirow}
\usepackage{threeparttable}

\AtBeginDocument{%
  \providecommand\BibTeX{{%
    \normalfont B\kern-0.5em{\scshape i\kern-0.25em b}\kern-0.8em\TeX}}}

\copyrightyear{2022}
\acmYear{2022}
\setcopyright{acmcopyright}
\acmConference[MM '22] {Proceedings of the 30th ACM International Conference on Multimedia }{October 10--14, 2022}{Lisbon, Portugal.}
\acmBooktitle{Proceedings of the 30th ACM International Conference on Multimedia (MM '22), October 10--14, 2022, Lisbon, Portugal}
\acmPrice{15.00}
\acmISBN{978-1-4503-9203-7/22/10}
\acmDOI{10.1145/3503161.3548303}
\acmSubmissionID{mmfp2391}

\begin{document}

\title{DuetFace: Collaborative Privacy-Preserving Face Recognition via Channel Splitting in the Frequency Domain}


\author{Yuxi Mi}
\email{yxmi20@fudan.edu.cn}
\orcid{0000-0002-1006-6041}
\affiliation{
  \institution{Fudan University}
  \city{Shanghai}
  \country{China}
}

\author{Yuge Huang}
\email{yugehuang@tencent.com}
\orcid{0000-0001-5387-5992}
\affiliation{
  \institution{Tencent Youtu Lab}
  \city{Shanghai}
  \country{China}
}

\author{Jiazhen Ji}
\email{royji@tencent.com}
\orcid{0000-0003-2708-9319}
\affiliation{
  \institution{Tencent Youtu Lab}
  \city{Shanghai}
  \country{China}
}

\author{Hongquan Liu}
\email{hqliu21@m.fudan.edu.cn}
\orcid{0000-0003-2544-4452}
\affiliation{
  \institution{Fudan University}
  \city{Shanghai}
  \country{China}
}

\author{Xingkun Xu}
\email{xingkunxu@tencent.com}
\orcid{0000-0001-6399-3415}
\affiliation{
  \institution{Tencent Youtu Lab}
  \city{Shanghai}
  \country{China}
}

\author{Shouhong Ding}
\email{ericshding@tencent.com}
\orcid{0000-0002-3175-3553}
\affiliation{
  \institution{Tencent Youtu Lab}
  \city{Shanghai}
  \country{China}
}

\author{Shuigeng Zhou} 
\authornote{Corresponding author: Shuigeng Zhou, School of Computer Science, and Shanghai Key Lab of Intelligent Information Processing, Fudan University.}
\email{sgzhou@fudan.edu.cn}
\orcid{0000-0002-1949-2768}
\affiliation{
  \institution{Fudan University}
  \city{Shanghai}
  \country{China}
}


\begin{abstract}
With the wide application of face recognition systems, there is rising concern that original face images could be exposed to malicious intents and consequently cause personal privacy breaches. This paper presents DuetFace, a novel privacy-preserving face recognition method that employs collaborative inference in the frequency domain. Starting from a counterintuitive discovery that face recognition can achieve surprisingly good performance with only visually indistinguishable high-frequency channels, this method designs a credible split of frequency channels by their cruciality for visualization and operates the server-side model on non-crucial channels. However, the model degrades in its attention to facial features due to the missing visual information.
To compensate, the method introduces a plug-in interactive block to allow attention transfer from the client-side by producing a feature mask. The mask is further refined by deriving and overlaying a facial region of interest (ROI). Extensive experiments on multiple datasets validate the effectiveness of the proposed method in protecting face images from undesired visual inspection, reconstruction, and identification while maintaining high task availability and performance. Results show that the proposed method achieves a comparable recognition accuracy and computation cost to the unprotected ArcFace and outperforms the state-of-the-art privacy-preserving methods. The source code is available at \url{https://github.com/Tencent/TFace/tree/master/recognition/tasks/duetface}.
\end{abstract}




\maketitle

\section{Introduction}

\emph{Face recognition} (FR) has been a phenomenal biometric method for identity authentication, with lots of remarkable breakthroughs gained in recent years. As face recognition is widely incorporated into applications in such as finance, health, and public security, there rises a growing concern about the privacy of sensitive facial images. The \emph{privacy-preserving face recognition} (PPFR) technique thus has arrested the attention of academia and the industry.

In a typical privacy-preserving face recognition scenario, a query face image is collected and held by local devices such as cell phones and webcams. As often constrained by computation power, they outsource the recognition task to a third-party service provider, who infers on a SOTA model pre-trained on massive labeled datasets. For clarity, we thereafter refer to the image holder as the client and the service provider as the server. As the query face here is considered private, the client is not willing to share the raw image with others, including the server.

In the recent decade, vast advancements have been incorporated into PPFR applications. We coarsely categorize them into two branches: the encrypt-based methods build the privacy of the face image on the bricks of cryptographic primitives and security protocols. Privacy is well protected as long as these bricks are provably or computationally secure. These methods, however, often bear prohibitive computation and communication costs. On the other hand, the transform-based methods perturb or regenerate the face image into a new representation that is visually indistinguishable from the server's perspective. The drawback of these methods is the almost inevitable participation of noise or loss of information, which results in a downgrade of recognition accuracy. Prior transform-based arts, therefore, generally face a paradoxical trade-off between privacy and accuracy.


\begin{figure}[tbp]
  \centering
  \includegraphics[width=\linewidth]{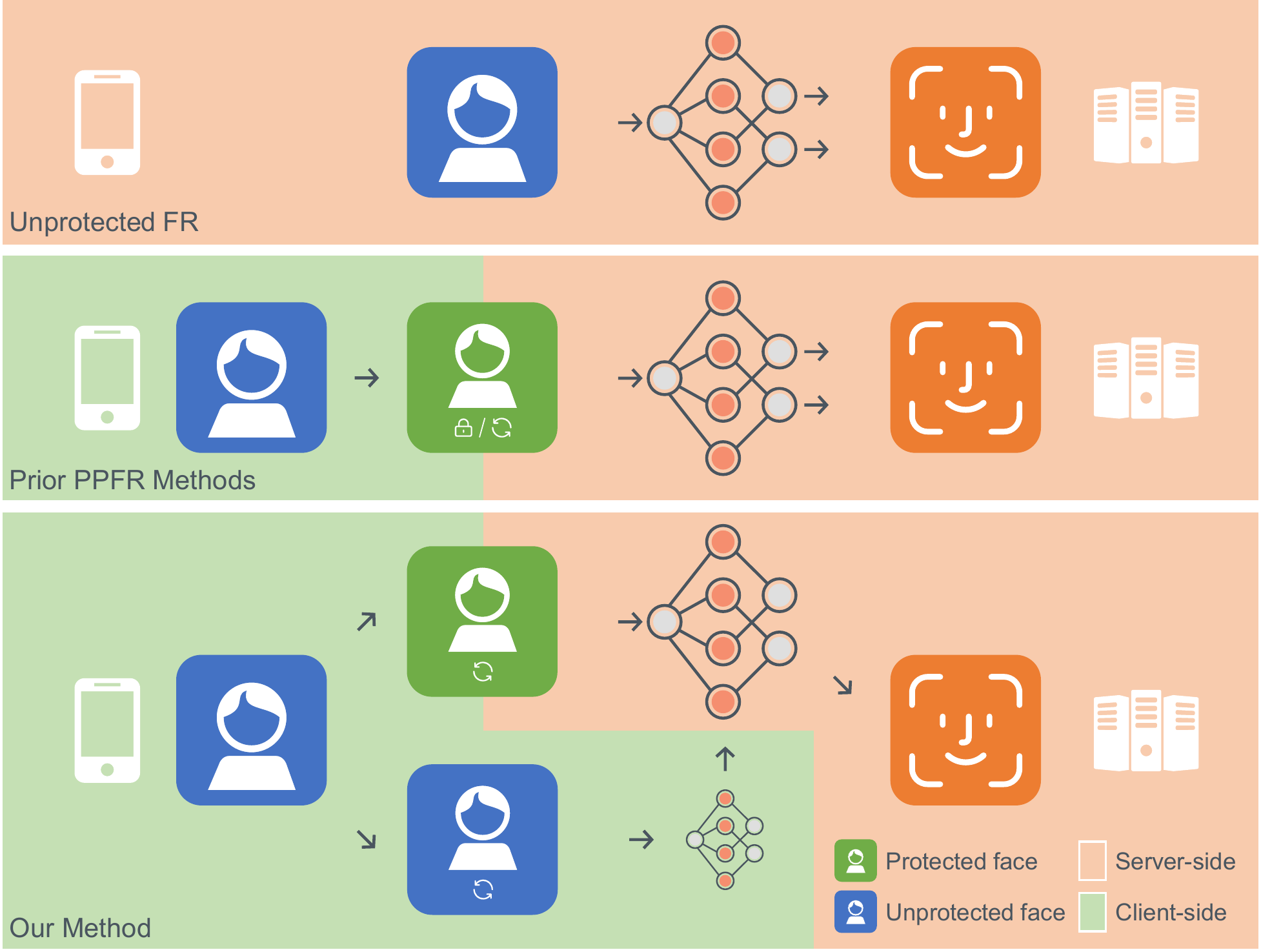}
  \caption{A paradigm comparison among unprotected FR (top), prior PPFR works (middle) and our method (bottom). The images are directly shared in unprotected FR, while are encrypted or transformed in PPFR. The server in prior works performs recognition alone, which usually suffers from a downgrade in accuracy. In our method, the recognition is performed through the collaboration of the server and client. }
  \Description{The comparison of paradigm among FR, prior PPFR works, and our method.}
  \label{fig:paradigm}
\end{figure}

Processing images in the frequency domain, first practically employed in the JPEG standard, is a time-honored approach for image compression. As a byproduct, prior researches observe that a small number of low-frequency channels aggregate most of the visual features that are believed crucial for recognition. We advance this observation with a counterintuitive discovery: \emph{the recognition can achieve a surprisingly acceptable accuracy even upon the removal of these crucial channels, while with far less visual information revealed}. We exploit this discovery as a starting point for our work.


Our work employs a transform-based approach as well. Yet, we propose a novel way to effectively solve the privacy-accuracy paradox by addressing them successively through a collaboration between the server and the client. Concretely, let the server possess a SOTA model, whereas the client holds a lightweight model that is an order of magnitude smaller in complexity and cost. We first \emph{split the frequency channels} into two parts by their cruciality for visualization, and let the server infer the non-crucial part. The visual information is therefore concealed from the server and, based on our novel discovery, at the price of tolerable performance degradation. Subsequently, we improve the accuracy by allowing the client to correct the server-side inaccurate attention on facial features, concretely, by \emph{transferring the client-side attention} to the server through a proposed interactive block. For a better understanding of our idea, a paradigm comparison among unprotected FR, prior PPFR works, and our method is illustrated in Fig.~\ref{fig:paradigm}. Extensive experiments validate the effectiveness of our scheme.

In summary, the contributions of our paper are three-fold:

\begin{enumerate}
    \item We propose a novel face recognition paradigm by combining the collaborative efforts of two parties: the client and the server. We develop a novel PPFR framework, referred to as DuetFace, which is satisfactory in recognition accuracy, good in cost-efficiency, and reliable in privacy protection;
    \item We introduce a channel splitting scheme to derive appropriate image break-ups and devise an interactive block based on ROI-refined feature masks to allow attention transfer;
	\item We conduct extensive experiments on multiple datasets, which validate the effectiveness and superiority of the proposed method.
\end{enumerate}


\section{Related Work}\label{sec:related-work}

\subsection{Privacy-Preserving Face Recognition}\label{sec:ppfr}

Significant advances of privacy-preserving face recognition (PPFR) has been achieved significant progress in the past decade, which can be roughly divided into two categories:

\noindent \textbf{Encrypt-based methods.} In this branch of work, face recognition is carried out on encrypted domains. Necessary computations such as feature extraction and similarity calculation are either accomplished straightly on the encrypted images or by executing certain security protocols. Pioneering works \cite{DBLP:conf/icisc/SadeghiSW09,DBLP:conf/pet/ErkinFGKLT09,DBLP:conf/icml/Huang0LA20} apply homomorphic encryption (HE) and garbled circuits (GC) to the Eigenface recognition algorithm to hide raw image features from undesired parties. Methods with similar intentions also employ other cryptographic primitives including matrix encryption \cite{DBLP:conf/acmturc/KouZZL21}, one-time-pad \cite{DBLP:conf/apccas/Ergun14a}, and functional encryption \cite{DBLP:conf/pkc/AbdallaBCP15}. Dedicated secure multiparty computation (MPC) schemes are introduced in \cite{DBLP:journals/iotj/MaLLMR19,DBLP:conf/ithings/YangZLLL18,DBLP:journals/soco/XiangTCX16} to perform certain operations (\emph{e.g.} parameter comparison) in a protected manner, or to outsource parts of the job to a trusted third party.

Encrypt-based methods have very little degradation on the recognition accuracy since almost all the operations involved are lossless. Their effectiveness is also strongly guaranteed by the provable or computational security of the employed cryptographic primitives. However, the practical usages of these methods are limited as they mostly bear intolerable computation costs and communication overhead. Moreover, these methods have low generalizability since most of them are tightly coupled with very specific face recognition schemes. Our work is more generalizable and can achieve competitive accuracy to the standard ArcFace with much less cost.

\noindent \textbf{Transform-based methods.} Another active line of research transforms face images into perturbed or regenerated representations to reduce their distinguishability from untrusted parties. For perturbation, differential privacy is employed by lots of works \cite{DBLP:journals/finr/ZhangHXGY20,DBLP:journals/compsec/ChamikaraBKLC20,DBLP:journals/ijon/LiWL19b,DBLP:conf/hotedge/X18} where raw images are overlayed by noise mechanisms to make them less visually differentiable. To increase anonymity, Honda et al. \cite{DBLP:conf/socpar/HondaOUN15} proposed a clustering-based method by mapping raw images to their class representations. As for means to regenerate images into new representations, some works exploit deep-learning-based methods such as adversarial generative network (GAN) \cite{DBLP:conf/btas/MirjaliliRR18,DBLP:journals/ijon/LiWL19b} and autoencoder \cite{DBLP:conf/icb/MirjaliliRNR18}, some alter facial landmarks \cite{DBLP:conf/icb/MirjaliliR17}, some employ channel shuffling \cite{wang22ppfrfd}, and some remove redundant information by mapping discriminative components into subspaces \cite{DBLP:conf/autoid/KevenaarSVAZ05,DBLP:conf/mlsp/ChanyaswadCMK16,DBLP:conf/www/MireshghallahTJ21}.

These methods usually face a trade-off in balancing privacy and accuracy since the means of protecting the images are performed, in essence, either by bringing in noise or discarding features, which bring in information loss with high probability. Our proposed method overcomes the trade-off and reduces the accuracy loss to a minimum level. 

\begin{figure*}
  \centering
  \includegraphics[width=0.98\linewidth]{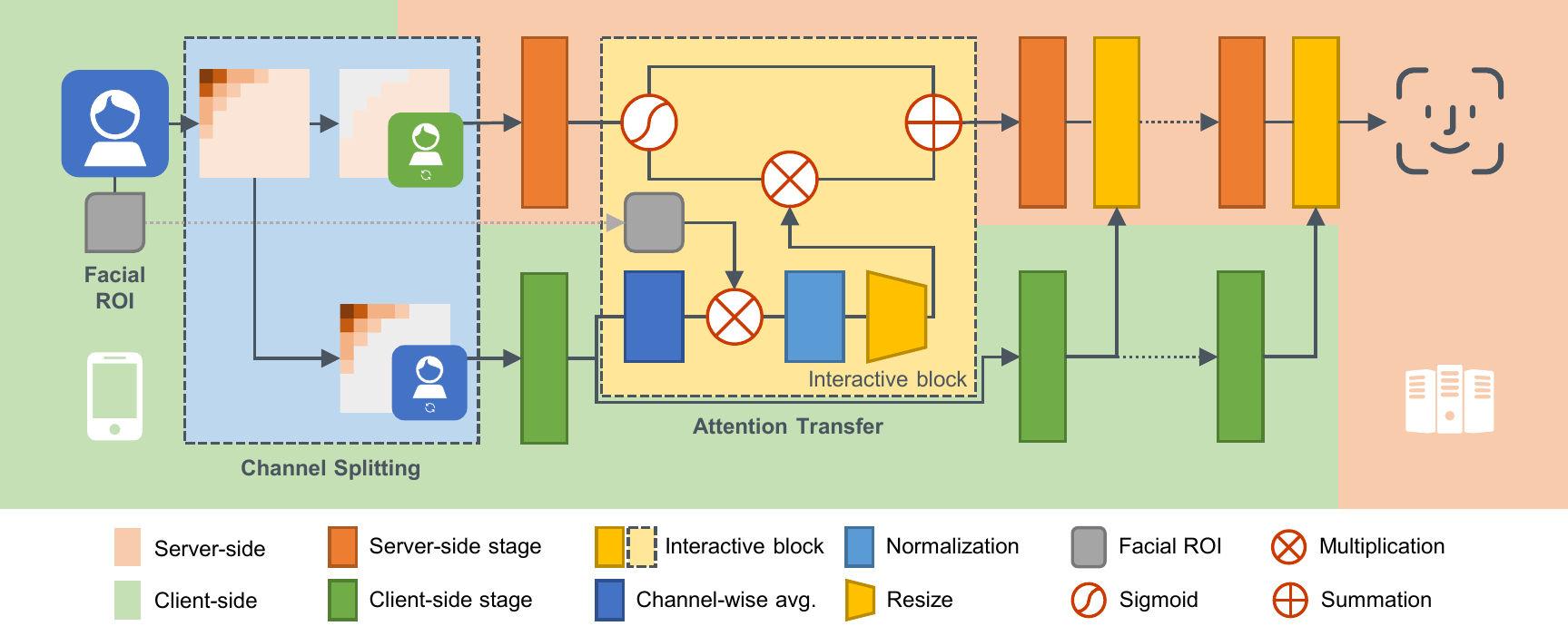}
  \caption{Architecture of DuetFace. The query image is split by the energy of its frequency channels into two parts and inferred separately by the client and the server. To compensate for accuracy loss, an interactive block is plugged in at the end of each stage, where a feature mask transfers the attention from the client to the server. A facial ROI is employed to refine the mask.}
  \Description{The architecture of DuetFace.}
  \label{fig:pipeline}
\end{figure*}

The recently proposed PPFR-FD \cite{wang22ppfrfd} is related to ours as both works study the contribution of channels on visualization and recognition. However, our findings differ from it in two aspects: (1) PPFR-FD begins with the premise that the lowest frequency channel contributes not much to the recognition task, which is quite different from ours; (2) The high-frequency channels are removed in PPFR-FD as they are believed to contribute little to distinguishability. These channels are instead retained and fully exploited in our work. Further, PPFR-FD realizes privacy mainly by randomizing the order of channels, while ours by removing visual components.


\subsection{Learning in the Frequency Domain}
Learning in the frequency domain is traditionally leveraged for image compression, which allows retaining meaningful patterns for image understanding tasks through compressed representations. Prior arts in face recognition \cite{DBLP:conf/iclr/TorfasonMATTG18,DBLP:conf/eccv/XuZR18} train autoencoder-based networks to perform compression and inference tasks simultaneously. \cite{DBLP:conf/ciarp/SantosA21} first performs image classification in the frequency domain directly. \cite{DBLP:conf/cvpr/0007QSWCR20} proposes an accuracy-retaining image down-sampling method, where spatial images are reorganized in the frequency domain to remove the non-crucial channels. 

\section{Methodology}\label{sec:methodology}

\subsection{Overview}\label{sec:overview} 

We here describe the proposed privacy-preserving face recognition method, referred to as DuetFace. In the world of art, a duet is a performance by two singers, instrumentalists, or dancers. Similarly, in DuetFace, the inference is carried out together by two parties, \emph{i.e.}, the server and the client.

\begin{figure}[tbp]
  \centering
  \includegraphics[width=\linewidth]{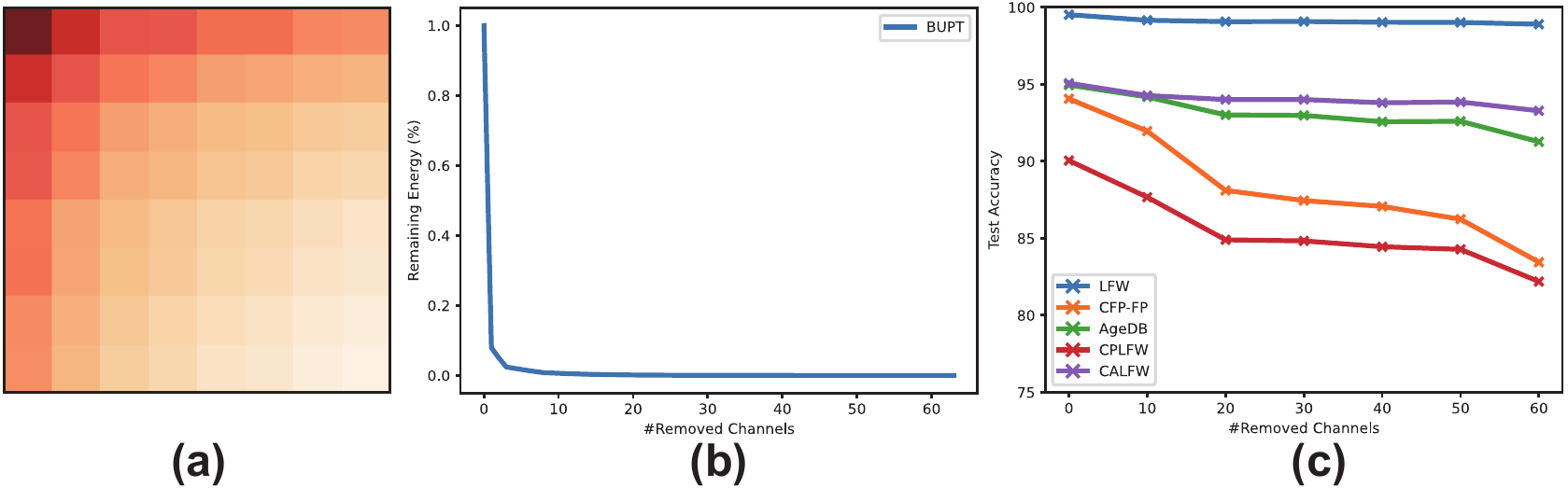}
  \caption{The contributions of channels to visualization and recognition. (a) The low-frequency channels on the top-left account for most of the visual information, in terms of channel energy. (b) The remaining energy sharply decreases as the low-frequency channels are discarded. (c) However, our experiments show that the models are able to maintain decent accuracy even with most visual information removed.}
  \Description{Channel splitting.}
  \label{fig:channel_splitting}
\end{figure}

\noindent \textbf{Our discovery and motivation.} 
Our work starts from a counter-intuitive discovery. Previous research in the frequency domain~\cite{DBLP:conf/cvpr/WangWHX20} suggests that the recognition is mainly determined by channels with lower frequency, which in the meantime are those with larger amplitude~\cite{DBLP:conf/cvpr/0007QSWCR20}, as they contribute most of the visual information. Yet, those prior arts mostly ignore the value of the rest ``non-crucial'' channels. We, on the contrary, surprisingly find that the models can also perform recognition with quite acceptable accuracy by using \emph{only the visually indistinguishable high-frequency channels}. To illustrate this, we train multiple recognition models, each with different numbers of lowest-frequency channels discarded in every color component, then evaluate the trained models on 5 public datasets. Results in Fig.~\ref{fig:channel_splitting}(b)(c) show that the models are able to maintain a quite decent performance even the remaining channels possess <10\% energy, although there certainly exists an accuracy gap due to the lack of visual information. This discovery allows us to construct our PPFR method from a completely different view. 

Concretely, we introduce a collaborative paradigm between the server and the client. We first design an appropriate split of the query image in the frequency domain, and let the server train on the high-frequency components in a privacy-preserving manner with tolerable accuracy loss. Then, we let the client further refine the server-side performance by compensating for the missing information without revealing the image itself. 

\noindent \textbf{The ability of the server and the client.} To better explain the motivation of our method, we start by characterizing the parties. We assume the server to be semi-honest and the client to be resource-sensitive. A semi-honest server is one who honestly follows the face recognition protocol and provides correct results while trying to learn as much as possible from the messages sent by the client. The server is an abstraction of corrupted or unregulated service providers, who may collect, use, and redistribute face images unauthorizedly. A resource-sensitive client is one bounded by limited storage, bandwidth, and computation power. The real-world clients are often reified as personal devices such as cell phones and webcams, whose owners may be unwilling to download and store large models locally or perform complex inference tasks.

\noindent \textbf{The security goals of DuetFace.} We address the inference-time privacy between the semi-honest server $\mathcal{S}$ and the client $\mathcal{C}$. The client possesses a query face image $X$ that it wants the server to identify. The image is considered private. We denote all the information related to $X$ that the client could share without causing a privacy breach as $I(X)$. Our privacy consideration is to prevent the unauthorized collection, use, and redistribution of $X$, which we concretize into three security goals:

\begin{enumerate}
    \item 	\textbf{Visual privacy.} As the very basis, the server should be unable to collect useful information from the visual appearance of the face image $X$;
    \item 	\textbf{Privacy against reconstruction.} The server may try to reconstruct the information it misses about $X$. We ergo prescribe, by leveraging $I(X)$, the server cannot effectively produce a reconstruction $X'$ of $X$;
    \item 	\textbf{Privacy against identity inference.} The server could redistribute the reconstructed image to a third party. Plus, the message could be intercepted during transmission. To prevent potential privacy leakage, we require that, without accessing the recognition model, acquiring either $I(X)$ or $X'$ should be insufficient to infer the identity of $X$.
\end{enumerate}

\noindent \textbf{The paradigm of DuetFace.} Let the server and the client each hold a local model $M_s$ and $M_c$, respectively. Here, $M_s$ is a full-size state-of-the-art model used to answer the recognition requests, whereas $M_c$ is a lightweight model applied as an aid. Both models are pre-trained by the server and $M_c$ is downloaded by the client. To infer a query image $X$, the client first splits $X$ into two appropriate parts $X_s$ and $X_c$, where visual information is removed from $X_s$ but retained in $X_c$. The client shares $X_s$ with the server and keeps $X_c$ to itself. The server identifies $X_s$ via its model $M_s$. Note that the loss of visual information in $X_s$ will surely degrade the performance of the server-side model. To compensate for the information loss, the client infers the other part $X_c$ of the image $X$ on $M_c$ at a very low computation cost, to obtain a concise supplementary representation $R(X_c)$, which will be revealed to the server. Finally, the server leverages $R(X_c)$ to refine its judgment and produce better results.

Fig.~\ref{fig:pipeline} shows the architecture of DuetFace, which consists of three major components: Channel splitting, attention transfer, and mask denoising. DuetFace is cost-efficient and fulfills the above-mentioned security goals at decent accuracy.

\subsection{Splitting Channels in Frequency Domain}\label{sec:splitting}

As the very first step of our proposed method, the image $X$ is split by channel frequency. 
To transform the face image to the frequency domain, we first follow the common data pre-processing protocol in the spatial domain to crop, resize and horizontally flip the face image, and obtain an input shape of $H$×$W$×3. Then, we convert the image from RGB to YCbCr color space and subsequently to the frequency domain by carrying out the block discrete cosine transform (BDCT) following the same way in JPEG compression \cite{DBLP:journals/cacm/Wallace91}. We also perform an 8-fold bilinear up-sampling right before BDCT. As standard BDCT maps each 8×8 pixel block into one frequency channel (here, maps 8$H$×8$W$×3 to $H$×$W$×192), the up-sampling enables us to maintain the channel shape and minimize the modifications to the recognition backbone.

We construct a credible split $\{X_s,X_c\}$ of $X$ by the amplitude of channels. Here, we measure amplitude by channel energy: given a channel, its energy is the mean of absolute values of all its elements. We first split channels by energy on the luma (Y) component of YCbCr color space, as it carries most of the visual profiles and features \cite{DBLP:conf/cvpr/0007QSWCR20}. As shown in Fig.~\ref{fig:channel_splitting}(a)(b), the low-frequency channels on the top-left corner contribute to >90\% of total energy. We select $K$ channels with the highest energy as the \emph{crucial channels} and regard the rest as \emph{non-crucial ones}. To achieve spatial consistency, the same selection is also applied to the chroma (Cb, Cr) components. Then, we produce $\{X_s,X_c\}$ utilizing the split channels. We form $X_c$ by concatenating the crucial channels (in the shape of $H$×$W$×3$K$), and $X_s$ by the rest. Therefore, visual information is convincingly removed from $X_s$ but retained in $X_c$, as illustrated in Sec.~\ref{subsec:visualization}.

We adjust the input shapes of the models $M_s$ and $M_c$ to meet the shapes of $X_s$ and $X_c$, respectively. It is widely observed that discarding non-crucial frequency channels affects the model very slightly \cite{DBLP:conf/cvpr/WangWHX20,DBLP:conf/cvpr/0007QSWCR20,wang22ppfrfd}, so we expect no performance change on $M_c$. As of $M_s$, results in Sec.~\ref{subsec:ablation} show that the model suffers a tolerable accuracy gap in the absence of visual information, which we are to compensate for in the following subsections.

\subsection{Attention Transfer}

\begin{figure}[tbp]
  \centering
  \includegraphics[width=\linewidth]{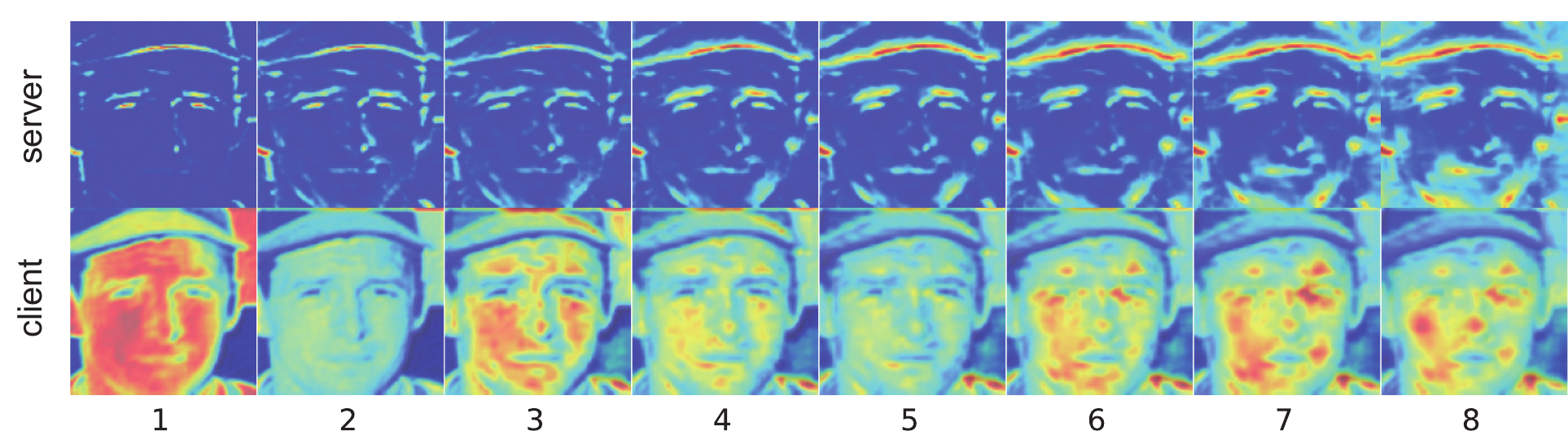}
  \caption{Visualization of each model layer via Grad-CAM. The server-side attention is inaccurate on facial features, while the accurate attention is retained by the client.}
  \Description{Visualization via Grad-CAM.}
  \label{fig:gradcam}
\end{figure}

We ascribe the downgrade of $M_s$ to the inaccuracy of model attention. To explain, we visualize the top-down attention maps for each layer of $M_s$ and $M_c$ via Grad-CAM~\cite{DBLP:journals/ijcv/SelvarajuCDVPB20}. As shown in Fig.~\ref{fig:gradcam}, after discarding the crucial channels, the attention of the server-side model $M_s$ could not focus on the effective visual features such as facial contours and the positions of eyes, noses, and lips, which are generally considered the indispensable information for high-quality face recognition. On the other hand, we observe that accurate attention is retained by the client-side $M_c$. To restore the missing information in $M_s$, we propose an \emph{interactive block} $IB$ that teaches the server-side $M_s$ about the attention on the client-side.

Currently, many popular face recognition backbones employ a stage-by-stage architecture to learn features of different levels. Assume $M_s$ and $M_c$ each has $n$ stages, we plug in our interactive block at the end of each stage. Concretely, for face image $X$, denote the feature maps of $M_s, M_c$ at the end of stage $i$ as $F_i(X_s), F_i(X_c)$, respectively. When querying $X$, the client first infers $X_c$ on $M_c$ to produce the feature map $F_i(X_c)$, and calculates its channel-wise average to obtain a one-channel feature mask $R_i(X_c)$. The client then normalizes $R_i(X_c)$ to $[0,1]$ and transmits it to the server. After receiving $R_i(X_c)$, the server first resizes it to align with the height and width of $F_i(X_s)$. Then the server activates its own feature $F_i (X_s)$ by passing it through a sigmoid function and updates it as:
$$F_i^{'} (X_s )=w_i \times (F_i (X_s ) \odot R_i (X_c ))+F_i (X_s ),$$
where $\odot$ is element-wise multiplication and $w_i$ is a trainable weight that decides the impact of the feature map on the server-side feature.

Note that we use the word ``interactive'' to describe the function of the block. In practice, the client is not required to jointly compute with the server for each block in real-time. Instead, it just infers $M_c$ once to retrieve and calculate the feature masks of all the stages $R(X_c )=\{R_i (X_c ),\cdots ,R_n (X_c )\}$, and sends it together with $X_s$ to the server. This simplifies the process.

\begin{figure}[tbp]
  \centering
  \includegraphics[width=0.9\linewidth]{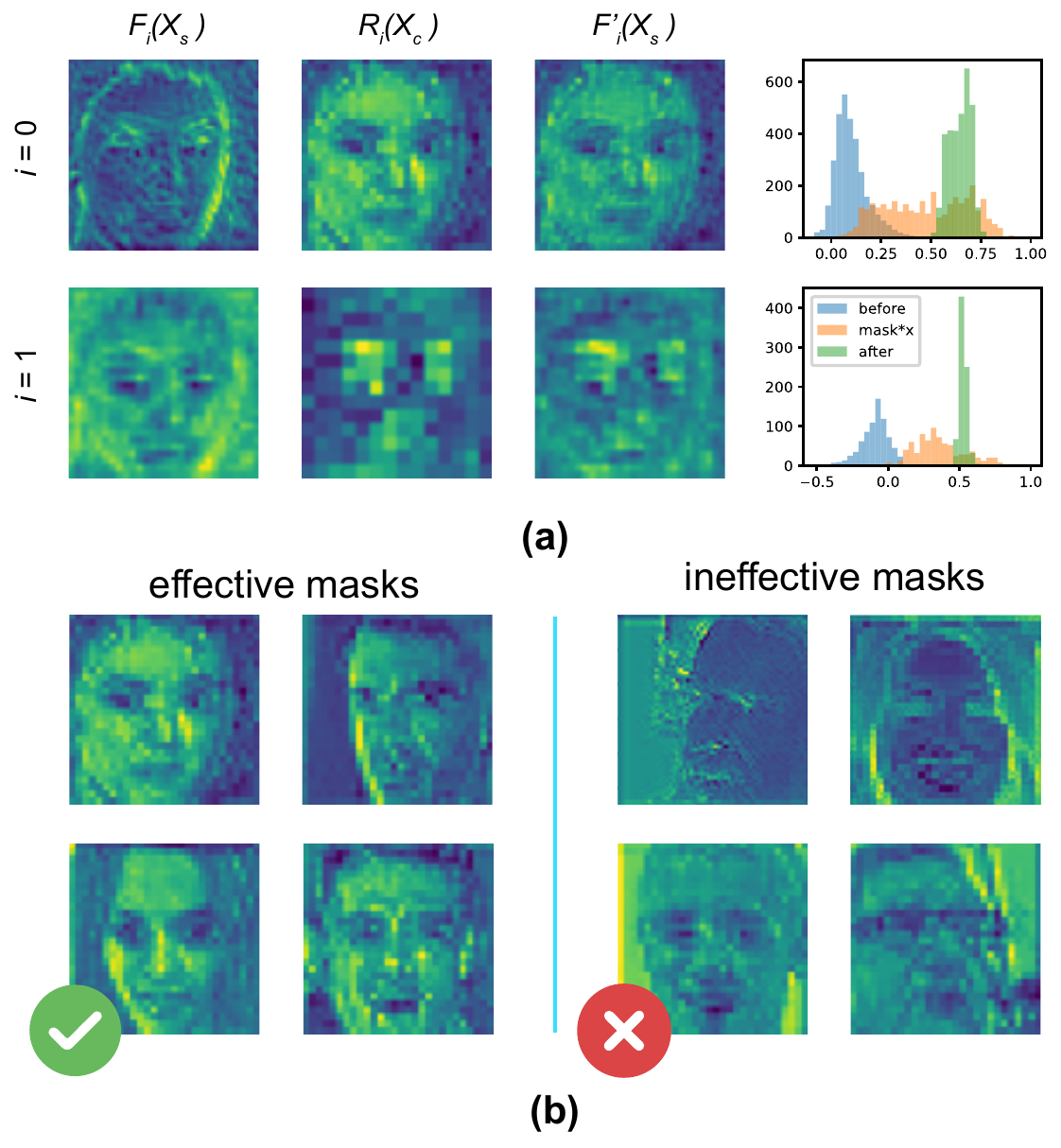}
  \caption{Illustration of attention transfer. (a) The mask $R_i(X_c)$ effectively transfers facial features from the client to the server for correcting the latter's inaccurate attention. (b) Effective masks highlight the facial contour correctly, while ineffective masks rather highlight the surroundings.}
  \Description{Illustration of attention transfer.}
  \label{fig:attention_transfer}
\end{figure}

Fig.~\ref{fig:attention_transfer}(a) illustrates attention transfer at stages 0 and 1 with samples of the server-side feature map before (left) and after (right) the interactive block, as well as the feature mask $R_i(X_c)$ (middle). The histograms show their corresponding value distributions. As inferred from both visualization and data distribution, the mask effectively transfers knowledge of features such as the entire facial contour (in stage 0) and the approximate positions of eyes and lips (in stage 1) to the server-side. The attention transfer leads to performance improvement of $M_s$, as demonstrated in Sec.~\ref{subsec:ablation}. To further improve the precision of the feature mask, we also propose a fast and effective denoising method detailed in the following subsection.

\subsection{Denoising the Mask by Facial ROI}

We subsequently refine the feature mask by removing noisy features. To explain, Fig.~\ref{fig:attention_transfer}(b) illustrates some counter-examples that the feature mask can sometimes be ineffective as it incorrectly highlights the hair, the headgear, and the surrounding area, rather than the face itself. Although these features are inherent and could be harmless in the standard face recognition process, in our case, they bring in undesirable noise during attention transfer.

We denoise the mask by obtaining a region of interest (ROI) on the raw image. Facial landmark detection is a proven technique that detects and tracks key points in a human face and is widely adopted in applications such as augmented reality. A general facial landmark detector produces a sequence of points that mark the positions of main facial features between facial contours and eyebrows, which specifies the interesting regions of our feature mask. Therefore, we utilize the point sequence to derive the facial ROI.

Specifically, right before performing BDCT, we pass the image $X$ through a pre-trained facial landmark detector to obtain the sequence of landmark points $P$. Here, we employ an open-source PFLD model \cite{DBLP:journals/corr/abs-1902-10859}. Note that PFLD can be replaced by an arbitrary lightweight 2D landmark detector. Subsequently, we derive the facial ROI, which wraps the facial region, by calculating the convex hull $H(P)$ of $P$ using Delaunay triangulation. Features outside the ROI are considered useless for attention transfer. To remove them, we turn $H(P)$ into a bool mask by marking all the pixels inside as 1 and the rest as 0, and overlay it on its feature masks:
$$R_i^{'}(X_c )=H(P) \odot R_i(X_c).$$
\noindent Ergo, a clean mask is produced. Finally, we normalize $R_i^{'}(X_c)$ and use it as a replacement of $R_i(X_c)$.

\begin{figure}[tbp]
  \centering
  \includegraphics[width=0.9\linewidth]{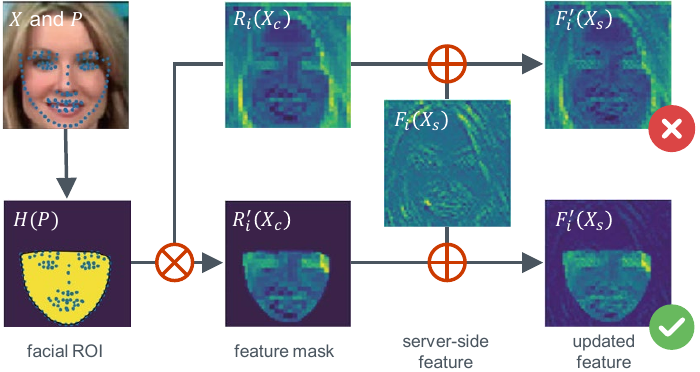}
  \caption{Effect of the facial ROI. After overlaying the ROI, the updated server-side feature shifts its highlight from the noisy surroundings (top) to the correct facial region (bottom).}
  \Description{Denoising the face image via facial ROI.}
  \label{fig:roi}
\end{figure}

Fig.~\ref{fig:roi} show sample masks before and after applying the facial ROI. It can be clearly observed that the ROI removes the features of the surroundings, resulting in focused attention to the face. Results in Sec.~\ref{subsec:ablation} also show a performance lift after employing the ROI.

As a brief summary, the proposed DuetFace successfully overcomes the privacy-accuracy trade-off problem. By performing channel splitting, privacy is guaranteed by removing visual information from the server’s side. Then, the accuracy is compensated by attention transfer from the client-side to the server-side via the ROI-refined feature mask. Sec.~\ref{subsec:sota} shows the effectiveness of our method, which almost fills in the accuracy gap with the standard ArcFace, and outperforms those SOTA privacy-preserving counterparts.

\section{Experiments}\label{sec:experiments}

\subsection{Datasets}
\noindent \textbf{Training datasets.} We employ MS1Mv2 as training set, plus BUPT-BalancedFace \cite{DBLP:journals/corr/abs-1911-10692} (BUPT in short) for ablation study. MS1Mv2 is refined from MS-Celeb-1M \cite{DBLP:conf/eccv/GuoZHHG16} and contains approximately 5.8M images of 85K individuals, whereas BUPT contains 1.8M images from 28K identities. For the training of facial landmark detector, we use Wider Facial Landmarks in-the-wild (WFLW) \cite{DBLP:conf/cvpr/Wu0YWC018} that contains 10K faces, each with 98 manual annotated landmarks.

\noindent \textbf{Evaluation datasets.} We benchmark our method on five popular datasets, including LFW \cite{LFWTech}, CFP-FP \cite{DBLP:conf/wacv/SenguptaCCPCJ16}, AgeDB \cite{DBLP:conf/cvpr/MoschoglouPSDKZ17}, CPLFW \cite{CPLFWTech}, and CALFW \cite{DBLP:journals/corr/abs-1708-08197}. LFW is the most commonly used evaluation dataset that contains 13K web-collected images from 5.7K identities. CFP-FP, AgeDB, CPLFW and CALFW embody a similar size while focusing on the performance on variations in such as pose and age. We also extend our evaluation to two general, large-scale benchmarks: IJB-B \cite{DBLP:conf/cvpr/WhitelamTBMAMKJ17} and IJB-C \cite{DBLP:conf/icb/MazeADKMO0NACG18}. 

\subsection{Implementation Details}

\noindent \textbf{Backbone.}
We use the adapted ResNet50 with an improved residual unit (IR-50) \cite{DBLP:conf/cvpr/HeZRS16} as the server-side backbone $M_s$, which has better convergence in the early training stages, and the MobileFaceNet \cite{DBLP:conf/ccbr/ChenLGH18} as the client-side $M_c$. For the client-side facial landmark detection, we apply a PFLD \cite{DBLP:journals/corr/abs-1902-10859} network. Note that both MobileFaceNet and PFLD are lightweight backbones dedicated to resource-constraint devices such as cell phones. We also adapt a smaller IR-18 backbone for ablation study, to explain the generality of our method on different network architectures.

\noindent \textbf{Preparation.}
We crop and resize each image to 112×112 pixels and add random flipping as image augmentation, then apply BDCT to obtain 192 frequency components by an open-source TorchJPEG library \cite{DBLP:conf/eccv/Ehrlich0LS20}. After transforming into the frequency domain, we calculate the energy, and select 10 channels from each of the Y, Cb, Cr components as stated in Sec.~\ref{sec:splitting}. This results in 30 channels selected in total. Note that we choose 30 without loss of generality and it is not the only choice. Therefrom, we form an $X_s$ in the shape of 112×112×162 and an $X_c$ of 112×112×30.

We alter the model input channels to meet the shapes of $X_c$ and $X_s$. Each of the IR-18/50 and MobileFaceNet models contains 4 stages. For attention transfer, we insert an interactive block $IB$ at the end of each stage. Aligning with the feature map shapes in $X_s$, in our case, the 4 feature masks are resized to the height and width of 56, 28, 14 and 7, respectively. The other parts of the models remain unchanged.

\noindent \textbf{Training.}
The IR-18/50 and MobileFaceNet models are trained for 24 epochs on the same dataset (either MS1Mv2 or BUPT), using the ArcFace \cite{DBLP:conf/cvpr/DengGXZ19} loss. We use the stochastic gradient descent (SGD) optimizer, which is applied with an initial learning rate of 0.1, a momentum of 0.9, and a weight decay of 5e-4, at a batch size of 512. We successively divide the learning rate by 10 at stages 10, 18, and 22. As of PFLD, we apply a learning rate and a weight decay of 1e-4 and 1e-6, respectively, and train it until convergence. Experiments are conducted on 8 NVIDIA Tesla V100 GPU under the PyTorch framework. The same random seed is sampled for all experiments for fairness.

\begin{table*}[tbp]
\caption{Comparisons with State-of-the-Art Methods}
\label{tab:sota}
\begin{tabular}{llccccccc}
\toprule
\textbf{Method}          & \textbf{PPFR} & \multicolumn{1}{l}{\textbf{LFW}} & \multicolumn{1}{l}{\textbf{CFP-FP}} & \multicolumn{1}{l}{\textbf{AgeDB}} & \multicolumn{1}{l}{\textbf{CPLFW}} & \multicolumn{1}{l}{\textbf{CALFW}} & \multicolumn{1}{l}{\textbf{IJB-B(TPR@FPR)}} & \multicolumn{1}{l}{\textbf{IJB-C(TPR@FPR)}} \\ \midrule
ArcFace \cite{DBLP:conf/cvpr/DengGXZ19}        & No   & 99.77                   & 98.30                      & 97.88                     & 92.77                     & 96.05                     & 94.13                     & 95.60                     \\
ArcFace-FD \cite{DBLP:conf/ciarp/SantosA21}     & No   & 99.78                   & 98.04                      & 98.10                     & 92.48                     & 96.03                     & 94.08                     & 95.64                     \\
\hline
PEEP \cite{DBLP:journals/compsec/ChamikaraBKLC20}            & Yes  & 98.41                   & 74.47                      & 87.47                     & 79.58                     & 90.06                     & 5.82                      & 6.02                      \\
Cloak \cite{DBLP:conf/www/MireshghallahTJ21}           & Yes  & 98.91                   & 87.97                      & 92.60                     & 83.43                     & 92.18                     & 33.58                     & 33.82                     \\
InstaHide \cite{DBLP:conf/icml/Huang0LA20}      & Yes  & 96.53                   & 83.20                      & 79.58                     & 81.03                     & 86.24                     & 61.88                     & 69.02                     \\
CPGAN \cite{DBLP:journals/tifs/TsengW20}          & Yes  & 98.87                   & 94.61                      & 96.98                     & 90.43                     & 94.79                     & 92.67                     & 94.31                         \\
PPFR-FD \cite{wang22ppfrfd}         & Yes  & 99.68                   & 95.04                      & 97.37                     & 90.78                     & 95.72                     & *                    & 94.10                          \\
\textbf{DuetFace (ours)} & \textbf{Yes}  & \textbf{99.82}          & \textbf{97.79}             & \textbf{97.93}            & \textbf{92.35}            & \textbf{96.10}            &  \textbf{93.66}                 &  \textbf{95.30}                    \\ \bottomrule

\end{tabular}
\begin{tablenotes}
\footnotesize
\item {* The results of PPFR-FD are quoted from [34] due to the lack of source code. Please note that its experimental condition may be different slightly from ours.}
\end{tablenotes}
\end{table*}

\subsection{Comparisons with SOTA Methods} \label{subsec:sota}

\noindent \textbf{Methods for comparison.} We compare DuetFace with two face recognition methods without privacy protection and five state-of-the-art PPFR methods. Specifically: (1) \textbf{ArcFace} \cite{DBLP:conf/cvpr/DengGXZ19} is the baseline method for RGB images without privacy protection; (2) \textbf{ArcFace-FD} is the same ArcFace backbone trained on frequency components instead, which we employ the method in \cite{DBLP:conf/ciarp/SantosA21}; (3) \textbf{PEEP} \cite{DBLP:journals/compsec/ChamikaraBKLC20} is the first practical method to introduce differential privacy in PPFR, where we set its privacy budget $\epsilon$ to 5; (4) \textbf{Cloak} \cite{DBLP:conf/www/MireshghallahTJ21} compresses the input feature space by a gradient-based perturbation. We set its accuracy-privacy parameter to 100; (5) \textbf{InstaHide} \cite{DBLP:conf/icml/Huang0LA20} is a lightweight encryption-based method in distributed setting that incorporates the mix-up of $k$ images, which we set to 2; (6) \textbf{CPGAN} \cite{DBLP:journals/tifs/TsengW20} generates compressive face representations by GAN and local differential privacy; And (7) \textbf{PPFR-FD} \cite{wang22ppfrfd} is a very recent method that masks the face by shuffling and mixing on frequency channels. Results are summarized in Tab.~\ref{tab:sota}.

\noindent \textbf{Results on LFW, CFP-FP, AgeDB, CPLFW, and CALFW} are reported by accuracy. 
Our method achieves a very close performance to the non-privacy-preserving ArcFace baseline. Concretely, our method achieves almost the same performance as ArcFace on LFW, AgeDB and CALFW, and has a limited accuracy drop of 0.51\% on CFP-FP, and 0.42\% on CPLFW. We believe such accuracy loss is because these two datasets possess more complex variations in pose, which downgrades the inference of facial landmarks. On the other hand, our method takes a leading role over all the SOTA privacy-preserving methods by an advantage from 0.14\% to 23.32\%.

\noindent \textbf{Results on IJB-B and IJB-C} are reported in TPR@FPR(1e-4), \textit{i.e.}, the true-positive rate at the false-positive rate of 1e-4. The TPR of DuetFace is slightly lower than ArcFace, but still outperforms most of its competitors, which indicates our method has good generalizability on large-scale datasets. We also notice that some SOTA methods may fail to generalize well as their performance degrades very sharply.

\subsection{Visual Privacy} \label{subsec:visualization}
We validate that our proposed method provides reliable privacy protection that satisfies the security goals stated in Sec.~\ref{sec:overview}. The basic concern among our goals is visual security, \textit{i.e.}, the server should not be able to visually inspect the query image by the information it acquires from the client. 

\begin{figure}[tbp]
  \centering
  \includegraphics[width=\linewidth]{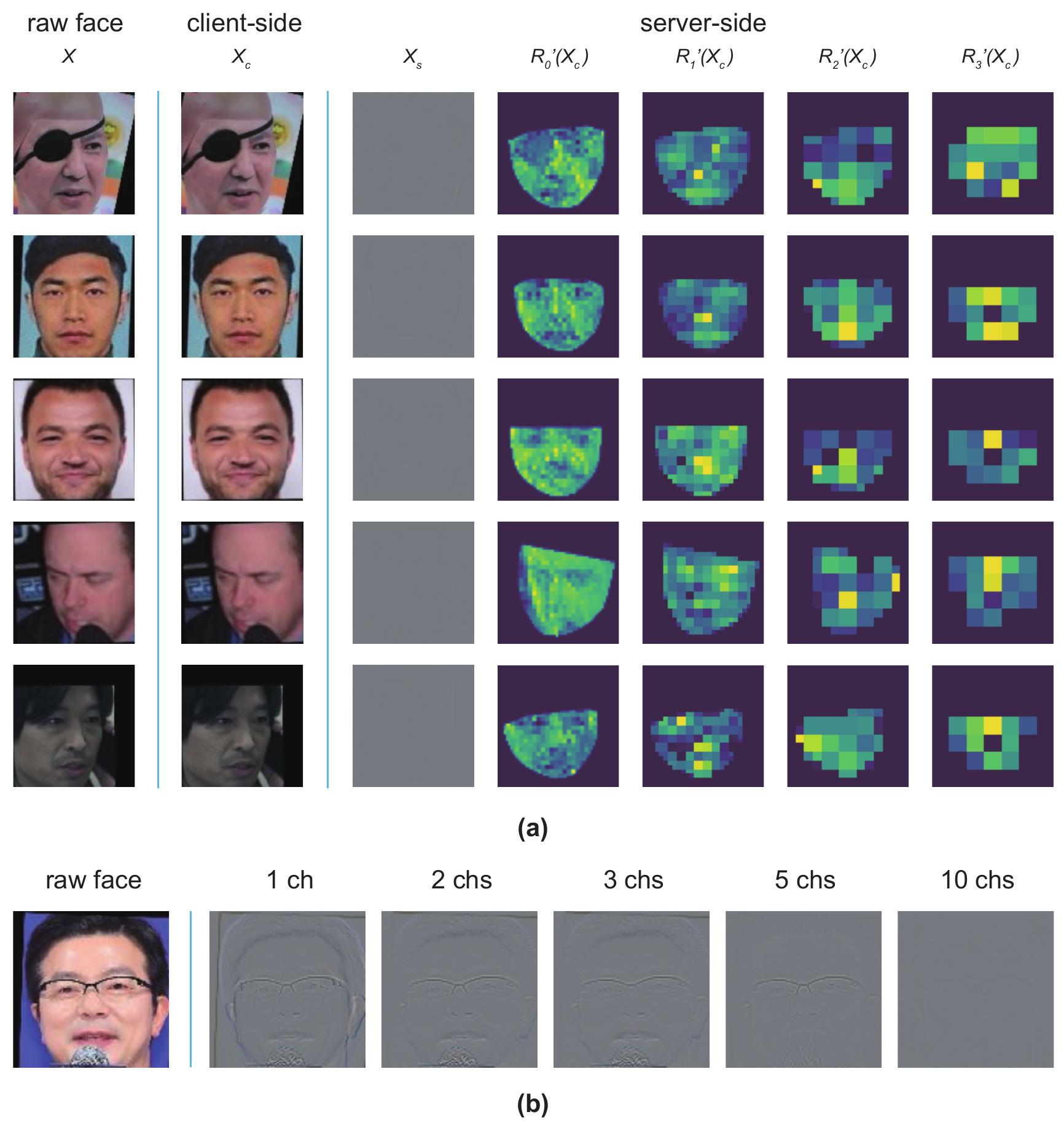}
  \caption{Visual privacy. (a) Illustration of $X_s$, $X_c$ and ${R_i^{'}(X_c)}$ on sample images. Perceptible visual information is removed from server-side. (b) We alter the number of crucial channels. As it increases, visual features become indistinguishable.}
  \Description{Visual privacy.}
  \label{fig:visualization}
\end{figure}

\noindent \textbf{Visualization of the client/server-side components and feature masks.} During the whole inference process, the information the server obtains including the non-crucial components $X_s$ of the query image and the refined feature masks ${R_i^{'}(X_c)}$ at each stage. To illustrate the effectiveness of our method, we randomly sample face images and visualize their $X_s$, $X_c$, and ${R_i^{'}(X_c)}$. As $X_s$ is in the frequency domain, we pad its removed channels with zero and convert it back to RGB by performing inverse DCT. We perform the same on $X_c$ for comparison. The results in Fig.~\ref{fig:visualization}(a) show that most of the perceptible visual information is removed from $X_s$ but retained in $X_c$, which prevents the server from inspecting $X_s$ while allowing the client to infer $X_c$ (in private) normally. As for the feature masks, the results show that they reveal only very limited visual information such as the approximate facial contours, which would not sabotage our security goal.

\noindent \textbf{Visualization on different numbers of crucial channels.} We alter the number of selected channels and visualize their corresponding $X_s$ in Fig.~\ref{fig:visualization}(b). Prior work \cite{wang22ppfrfd} also discards one channel with the highest energy (the direct current component) in its means to protect privacy. Our visualization implies such removal could be insufficient since part of the visual features can still be inferred from the rest components. As the number of selected channels increases, the remaining visual features become indistinguishable. 

\subsection{Privacy Against Malicious Intents} \label{subsec:reconstruction}

\begin{figure}[tbp]
  \centering
  \includegraphics[width=0.98\linewidth]{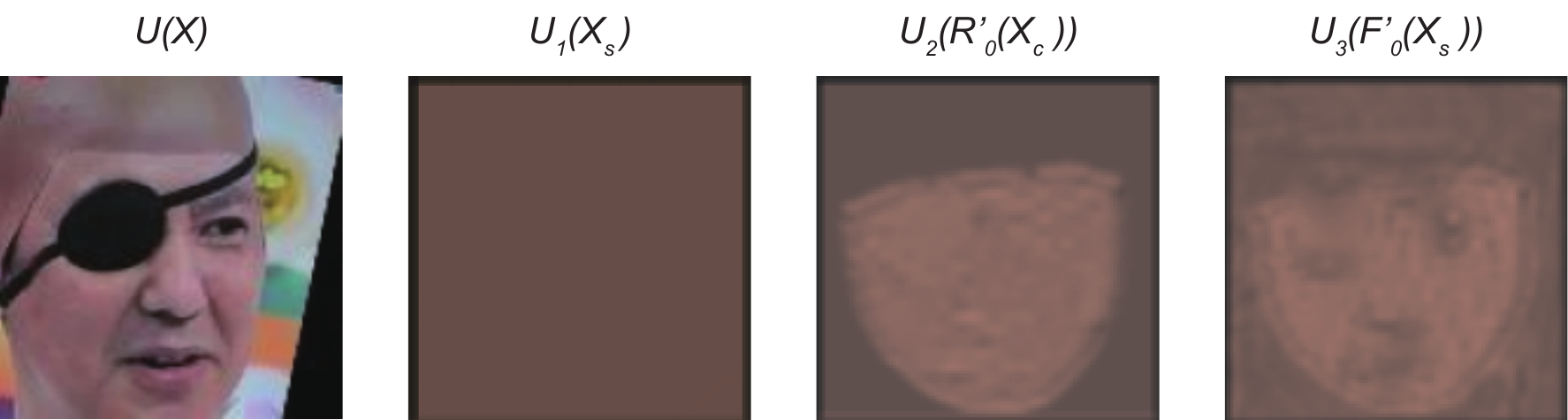}
  \caption{Reconstruction outcomes on $X_s$, $R_0^{'}(X_c)$ and $F_0^{'}(X_s)$ of sample image via U-Net. The server obtains hardly any visual feature. Reconstruction on the raw face (leftmost) is also provided as a validation of the ability of U-Net.}
  \Description{Reconstruction.}
  \label{fig:reconstruction}
\end{figure}

Ever since its invention, face recognition is under the threat of malicious attacks. Our security goal in Sec.~\ref{sec:overview} addresses concerns to two major forms of inference threats. Specifically, (1) though visually inaccessible, the semi-honest server may still convincingly reconstruct the query image based on what it possesses; And (2) any third party, if intercepts the information transferred from the client or receives unauthorized redistribution from the server, may infer the identity of the face even without accessing the recognition model. To further demonstrate our method's ability to protect valuable information about the query face, we impose these two types of threats on DuetFace. 

\begin{table}[htbp]
\caption{Effectiveness Against Reconstruction and Malicious Identity Inference}
\label{tab:reconstruction}
\begin{tabular}{llll}
\toprule
\textbf{Target}              & \textbf{SSIM↓}  & \textbf{PSNR↓} & \textbf{Accuracy↓} \\
\midrule
Raw image           & 0.9993 & 48.30 & 99.80          \\
$X_s$               & 0.5395 & 13.01 & 51.52          \\
Reconstruction of $X_s$      & 0.5012       & 12.89      &  52.03         \\
Reconstruction of $R_0^{'}(X_c)$             & 0.4422    &  12.45     & 57.18          \\
Reconstruction of $F_0^{'}(X_s)$             & 0.4555    &  12.63     & 61.97         \\
\bottomrule
\end{tabular}
\end{table}

\noindent \textbf{Effectiveness against reconstruction.} Autoencoders are widely used for reconstruction. In our case, the server may attempt to reconstruct the face image by an autoencoder-based network. Accessible information for the server including the non-crucial channels, the feature masks and their combination, which equals to the masked server-side feature map $F_i^{'}(X_s)$. Ergo, we first infer images randomly picked from MS1Mv2 on a pre-trained DuetFace to collect their $X_s$, $R_0^{'}(X_c)$ and $F_0^{'}(X_s)$. Note that in our case, as the resolution of feature maps is successively divided by half in each stage, the reconstruction is most likely to succeed on stage 0. We train three autoencoder-based U-Net models, denoted as $U_1$, $U_2$ and $U_3$, on $X_s$, $R_0^{'}(X_c)$ and $F_0^{'}(X_s)$, respectively, then use the trained model for reconstruction. We quantify the quality of the reconstructed images by structural similarity index (SSIM, as compared to the raw image) and peak signal-to-noise ratio (PSNR). As shown in Fig.~\ref{fig:reconstruction}, none of $U_1$, $U_2$ and $U_3$ manages to effectively reconstruct an image that embodies distinguishable features. And in Tab.~\ref{tab:reconstruction}, we can see that the SSIM and PSNR values are low for all three cases. These verify the robustness of our method against reconstruction.

\noindent \textbf{Effectiveness against identity inference.} We subsequently feed $X_s$ (after transformed into the spatial domain) and the reconstructed images into a pre-trained standard ArcFace model and see if they can be recognized. Results are summarized in the last column of Tab.~\ref{tab:reconstruction}. The intention to infer from $X_s$ or its reconstruction $U_1({X_s})$ fails, as the accuracy is slightly higher than random guess (50\%). A little accuracy gain is observed on the recognition of $U_2({R_0^{'}( X_c)})$ and $U_3({F_0^{'}(X_s)})$, as the mask could provide a trace amount of information. Yet, the accuracy remains very low, which thus does not incur an effective threat to our method.

\subsection{Ablation Study} \label{subsec:ablation}

\begin{table}[htbp]
\caption{Ablation Study Results}
\label{tab:ablation}
\begin{threeparttable}
\begin{tabular}{lccccc}
\toprule
\textbf{Method} & \multicolumn{1}{l}{\textbf{LFW}} & \multicolumn{1}{l}{\textbf{CFP-FP}} & \multicolumn{1}{l}{\textbf{AgeDB}} & \multicolumn{1}{l}{\textbf{CPLFW}} & \multicolumn{1}{l}{\textbf{CALFW}} \\
\midrule
\multicolumn{6}{l}{\textbf{IR-50 + MobileFaceNet, on MS1Mv2}}                                                                                                                                           \\
ArcFace \cite{DBLP:conf/cvpr/DengGXZ19}        & 99.77                            & 98.30                               & 97.88                              & 92.77                              & 96.05                              \\
DuetFace        & 99.82                            & 97.79                               & 97.93                              & 92.35                              & 96.10                              \\
w/o $IB$        & 99.70                            & 95.86                               & 97.57                              & 90.82                              & 95.86                              \\
w/o ROI         & 99.78                            & 97.23                               & 97.85                              & 92.07                              & 96.06                              \\
w/o $M_s$            & 99.48                            & 93.91                               & 96.10                              & 89.68                              & 95.08                              \\
\hline
\multicolumn{6}{l}{\textbf{IR-18 + MobileFaceNet, on BUPT}}                                                                                                                                             \\
ArcFace \cite{DBLP:conf/cvpr/DengGXZ19}        & 99.52                            & 94.06                               & 94.95                              & 90.05                              & 95.07                              \\
DuetFace        & 99.40                            & 93.79                               & 95.07                              & 89.67                              & 95.03                              \\
w/o $IB$        & 99.16                            & 91.96                               & 94.18                              & 87.67                              & 94.27                              \\
w/o ROI         & 99.39                            & 93.46                               & 95.12                              & 89.03                              & 94.86                              \\
w/o $M_s$            & 99.32                            & 92.43                               & 93.42                              & 89.15                              & 93.62                             \\
\bottomrule
\end{tabular}
\begin{tablenotes}
\footnotesize
\item[*] Two sets of experiments are performed in combination of IR-50+MS1Mv2 and IR-18+BUPT to demonstrate the generality of our method.
\item[**] ArcFace is the unprotected baseline. ``w/o $IB$'',  ``w/o ROI'', ``w/o $M_s$'' correspond to the removal of the interactive block, ROI and server-side model, respectively.
\end{tablenotes}
\end{threeparttable}
\end{table}

We analyze the effects of the major components of DuetFace by testing the performance when removing one of them. Results are reported in Tab.~\ref{tab:ablation} on LFW, CFP-FP, AgeDB, CPLFW, and CALFW. To validate the generality of our method, we also report the results on the combination of the IR-18 model and BUPT dataset.

\noindent \textbf{Effect of the interactive block.} The server leverages client-side features to compensate for its inaccurate attention to facial features. To demonstrate this, we remove the interactive block as well as the client-side model completely and train the server-side $M_s$ alone on $X_s$. Note that this setting is equal to ``w/o $M_c$''. From Tab.~\ref{tab:ablation}, we observe a <2.5\% accuracy degradation. 

\noindent \textbf{Effect of facial ROI.} The facial ROI refines the feature mask by focusing its attention on the facial region. We remove the facial landmark detector and let the client send the unrefined feature mask directly to the server. We see a notable performance downgrade, especially on CFP-FP and CPLFW. This is possibly because the variety of poses on these two datasets exacerbate attention shifting from the facial region to its surrounding area, which is undesirable.

\noindent \textbf{Performance of the client-side model alone.} The client-side model is implemented as an aid to the server-side. A large performance gap is observed between $M_c$ alone and DuetFace, as $M_c$ is much smaller in scale and complexity.

\subsection{Complexity and Cost} \label{subsec:efficiency}

\begin{table}[htbp]
\caption{Complexity and Cost of DuetFace}
\label{tab:efficiency}
\begin{tabular}{lllll}
\toprule
\textbf{Metric}                              & \textbf{ArcFace}    & \textbf{DuetFace}      & \textbf{$M_s$} & \textbf{$M_c$} \\
\midrule
Storage (\#param)        & 43.59M & 46.53M & 44.03M  & 2.50M   \\
Time (s/batch)            & 0.8492     & 2.2153     & 1.2165      & 0.9988      \\
Comm. (\#elements) & 37,632     & 35,525     & N/A         & N/A        \\ \bottomrule
\end{tabular}
\end{table}

To demonstrate the resource-friendliness of our method, we summarize the model size, time cost, and communication overhead required for inference in Tab.~\ref{tab:efficiency}. As the availability of our method mainly depends on the client-side budget, for a clear demonstration, we list the space and time costs of the client and server separately. The PFLD model is included when evaluating the client's cost. 

\noindent \textbf{Model size.} We list the model size by the number of parameters. Since both MobileFaceNet and PFLD are an order of magnitude smaller than the server-side model, the employment of the local model only accounts for a 6.7\% increase of the whole model size and takes about 10M of storage space at the client-side.

\noindent \textbf{Inference time.} We perform inference on images with a batch size of 64 and record the average per-batch time. Here, the inference is performed in an asynchronous manner, i.e., the client completes the local computations on all query images, then hands it over to the server for the rest. As compared to ArcFace, our total inference time increased by ×2.6. We think such an increase is marginal as it is the communication time that predominates in real-world practice. The overall time cost is still within a decent scope. 

\noindent \textbf{Communication overhead.} We calculate by adding up the number of elements in the server-side components $X_s$ and the feature masks of all 4 stages ${R_i^{'}(X_c)}$. The baseline ArcFace transfers the RGB images of 3×112×112 directly, resulting in 37,632 elements. In our original purpose, each image is up-sampled before BDCT (as means to preserve the same input height and width), which causes a larger communication cost. We can overcome this by moving the up-sampling later to the server-side. Therefore, the server-side components $X_s$ is of shape 14×14×160, containing 31,360 elements. On the other hand, the 4 feature masks are the shape of 56×56, 28×28, 14×14, and 7×7, respectively, which accounts for 4,165 elements in total. Thus, the total element number is 35,525, making our communication overhead similar to that of the baseline ArcFace.

\section{Conclusion}\label{sec:conclusion}
This paper presents DuetFace, a novel PPFR method that achieves high recognition performance through the collaboration between the client and the server. Channel splitting and attention transfer in the frequency domain are leveraged to implement the proposed method. Concretely, our method trains and infers the server-side model on visually indistinguishable non-crucial channels, and compensates for the inaccurate attention by the client-side information, in particular, by producing and transferring a feature mask through a plug-in interactive block. We refine the feature mask by overlaying a facial ROI. Extensive experiments show the proposed method is satisfactory in recognition accuracy, with good cost-efficiency, and achieves high reliability in privacy protection.

\bibliographystyle{ACM-Reference-Format}
\bibliography{DuetFace_bib}

\end{document}